# Klasifikasi Komponen Argumen Secara Otomatis pada Dokumen Teks berbentuk Esai Argumentatif

**DERWIN SUHARTONO**

**Abstrak**

Dengan pengenalan argumen secara otomatis dari dokumen teks, penulis esai dapat melakukan inspeksi pada teks yang mereka tulis. Hal ini akan membantu penilaian esai secara lebih objektif dan tepat karena penilai esai akan melihat seberapa baik komponen argumentasi terbentuk. Beberapa peneliti mencoba untuk melakukan pendeteksian dan klasifikasi argumen serta penerapannya pada berbagai domain. Mayoritas pendekatan yang digunakan adalah menggunakan feature extraction dari teks. Secara umum feature yang digunakan adalah structural, lexical, syntactic, indicator, dan contextual. Pada penelitian ini, diberikan feature tambahan pada sekumpulan feature yang sudah sering digunakan. Feature yang ditambahkan mengadopsi dari list keyword yang diberikan oleh Knott dan Dale (1993). Hasil dari eksperimen menunjukkan akurasi klasifikasi jenis argumen sekitar 72.45%. Selain itu, diperoleh juga bahwa tanpa penggunaan list keyword tersebut, akurasi yang dicapai masih pada nilai yang sama. Hal ini memberikan pandanganbahwa penggunaan list keyword tidak mempengaruhi signifikansi dari kualitas feature. Feature yang digunakan juga masih lemah untuk mengklasifikasikan major claim dan claim, sehingga butuh tambahan feature yang bisa menjadi representasi pengetahuan dari kedua jenis komponen argumen itu.

1. **Pendahuluan**

   Menulis esai merupakan bagian yang penting khususnya untuk hal yang terkait dengan pendidikan. Penugasan untuk menulis berbagai jenis esai sudah diberikan kepada siswa sejak duduk di bangku dasar. Dimulai dari bentuk esai paling sederhana yang berjumlah hanya 1 paragraf sampai kepada esai kompleks yang terdiri dari 2, 3 atau banyak halaman. Fakta ini memberikan pengertian bahwa menulis esai adalah hal yang esensial untuk bisa dikuasai dengan baik.

Dalam pelajaran bahasa khususnya mengenai esai, terdapat beberapa jenis esai, diantaranya adalah deskripsi, narasi, persuasi, eksposisi, dan argumentasi. Pembuatan esai memiliki pendekatan yang bervariasi tergantung pada setiap jenisnya. Selain cara membuat yang berbeda, cara untuk memberikan penilaiannya pun berbeda-beda. Secara umum, penilaian esai memiliki kesamaan pada bagian pengecekan grammar, spelling, dan stylistic. Akan tetapi, lebih spesifik lagi masih ada bagian yang harus dievaluasi sesuai dengan jenis esainya. Paper ini membahas secara khusus mengenai aspek yang perlu dilihat lebih jauh dari esai jenis argumentasi.

Esai argumentasi merupakan esai yang membutuhkan evaluasi lebih kompleks sehingga bisa dihasilkan skor yang lebih objektif. Dalam mengevaluasi esai argumentasi, salah satu bagian yang penting untuk diobservasi adalah bagaimana struktur argumentasi bisa dibangun menjadi kesatuan yang utuh. Apabila struktur tersebut berhasil dibuat dan dirangkai dengan baik, maka akan dihasilkan esai argumentasi yang berkualitas.

Menurut Peldszus dan Stede (2013), sebuah argumen terdiri dari beberapa komponen dan menunjukkan sebuah struktur yang didasari oleh hubungan argumentatif antar komponen. Stab dan Gurevych (2014) menjelaskan lebih lanjut bahwa komponen argumen memuat sebuah claim yang didukung atau ditolak setidaknya oleh satu premis. Claim adalah pusat dari komponen dalam sebuah argumen. Claim merupakan kalimat yang kontroversial yang seharusnya tidak bisa diterima oleh pembaca apabila tidak ada kalimat berikutnya yang mendukung claim tersebut. Premise merupakan komponen yang menggambarkan validitas dari claim.

Dengan pengenalan argumen secara otomatis dari dokumen teks, penulis esai dapat melakukan inspeksi pada teks yang mereka tulis seperti contohnya perbaikan struktur untuk peningkatan kualitas argumentasi. Asumsi ini didukung dengan penemuan pada bidang psikologi yang mengkonfirmasi bahwa sekalipun hanya tutorial umum yang digunakan, hal tersebut bisa meningkatkan kualitas dari argumen yang dituliskan (Butler dan Britt, 2011). Dan tentunya akan membantu penilaian esai secara lebih objektif dan tepat karena penilai esai akan melihat seberapa baik komponen argumentasi terbentuk.

Akhir-akhir ini beberapa peneliti mencoba untuk melakukan pendeteksian dan klasifikasi argumen serta penerapannya pada berbagai domain. Mayoritas pendekatan yang digunakan adalah menggunakan feature extraction dari teks, namun ada juga pendekatan rule-based. Moens, Boiy, Palau, dan Moens (2007) melakukan eksperimen dengan mendeteksi komponen argumen pada dokumen hukum. Madnani, Heilman, Tetreault dan Chodorow (2012) mengkombinasikan pendekatan rule-based dengan probabilistic sequence model untuk secara otomatis mendeteksi high-level organizational element pada wacana argumentatif. Ekstraksi argumen yang ada pada formulasi kebijakan publik dilakukan oleh Florou, Konstantopoulos, Kukurikos, dan Karampiperis (2013). Pendekatan berbeda dilakukan oleh Ong, Litman, dan Brusilovsky (2014) yang menggunakan ontology-based dalam pendeteksian komponen argumen serta mengujicobakan hasilnya pada penilaian esai secara otomatis. Dalam rangka menyediakan korpus yang bisa dijadikan referensi dalam pendeteksian komponen argument, Stab dan Gurevych (2014) melakukan anotasi manual pada 90 esai argumentasi. Pada tahun yang sama, Stab dan Gurevych (2014) menggunakan output korpus dari paper anotasi manualnya itu untuk melakukan pekerjaan pendeteksian komponen argumen serta hubungan antar komponen argumen. Di sisi lain, Song, Heilman, Klebanov, dan Deane (2014) mencoba melakukan penilaian pada esai argumentasi dengan menggunakan kumpulan feature yang kurang lebih sama dengan yang digunakan pada peneliti-peneliti sebelumnya.

Pada eksperimen kali ini, diberikan penambahan feature pada sekumpulan feature yang pernah digunakan oleh penelitian sebelumnya. Feature yang ditambahkan mengadopsi dari list keyword yang dimiliki oleh Knott dan Dale (1993). Pada bagian akhir dari paper ini diperlihatkan perbandingan hasil eksperimen tanpa menggunakan feature tambahan tersebut dan dengan menggunakannya.

2. **Kajian Literatur**

Pendeteksian komponen argumen dari satu dokumen teks sudah banyak dilakukan oleh para peneliti. Secara mayoritas,pendekatan yang digunakan adalah menggunakan pendekatan machine learning dengan memanfaatkan feature extraction. Beberapa peneliti juga mencoba untuk melakukan pendeteksian dengan menggunakan pendekatan rule-based. Lebih lanjut lagi, akhir-akhir ini setelah pendeteksian komponen argumen selesai dilakukan, peneliti langsung mencoba pemanfaatannya untuk proses penilaian esai secara otomatis.

Moens, Boiy, Palau, dan Reed (2007) melakukan penelitian untuk mendeteksi komponen argumen pada dokumenhukum. Pencarian komponen argumen dilakukan dengan melalui klasifikasi. Pada awalnya classifier diberikan pelatihan dengan menggunakan sekumpulan argumen yang sudah dianotasi. Feature set yang digunakan adalah properti lexical, semantic dan discourse dari teks. Feature tersebut diantaranya adalah unigram, bigram, trigram, adverb, verb, modal auxiliary, word couple, text statistic, punctuation, keywords, dan parse feature. Keyword yang digunakan pada feature tersebut diadopsi dari daftar kata-kata yang dihasilkan oleh Knott dan Dale (1993). Algoritma klasifikasi yang digunakan adalah multinomial naïve Bayes dan maximum entropy model. Berbagai kombinasi feature diujicobakan, dan hasil yang terbaik diperoleh dari kombinasi antara word couple (dipilih dari POS-tag-nya), verb, dan text statistic.

Pendekatan untuk mendeteksi komponen argumen yang berbeda dilakukan oleh Madnani, Heilman, Tetreault dan Chodorow (2012). Dengan memanfaatkan rule-based bersama dengan probabilistic sequence model, mereka mencoba untuk mengidentifikasi high-level organizational element dari sebuah wacana argumentatif. Organizational element yang dimaksud disebut juga sebagai shell language. Rule-based yang digunakan adalah dengan memanfaatkan sekumpulan 25 pola hand-written regular expression. Anotasi secara manual tanpa guideline yang baku dilakukan pada 170 esai oleh orang-orang yang berpengalaman dalam penulisan esai. Contoh dari pola pengenalan shell language bisa dilihat pada gambar 1. Sedangkan sequence model yang digunakan adalah berdasar pada conditional random field (CRF) dengan menggunakan sejumlah kecil general feature berdasar pada frekuensi leksikal.

```
MODAL → do | don't | can | cannot | will | would | ...
ADVERB → strongly | totally | fundamentally | vehemently | ...
AGREEVERB → disagree | agree | concur | ...
AUTHORNOUN → writer | author | speaker | ...
SHELL → I [MODAL] [ADVERB] AGREEVERB with the AUTHORNOUN
```

**Gambar 1. Contoh pola pengenalan shell language**
**(Madnani, Heilman, Tetreault dan Chodorow, 2012)**

Domain lain yang dicoba oleh para peneliti untuk ekstraksi argumen adalah dalam rangka mendukung formulasi kebijakan publik (Florou, Konstantopoulos, Kukurikos, dan Karampiperis, 2013). Penelitian ini bisa membantu pembuat kebijakan untuk melihat bagaimana reaksi yang diperoleh ketika kebijakan tersebut sudah benar-benar diberikan kepada masyarakat. Pada penelitian ini, tense dan mood digunakan sebagai indikator utama argumen. Feature yang digunakan terdiri dari 41 jenis feature yang terkategorikan menjadi 5 jenis yaitu:

- Jumlah kemunculan discourse marker dari sebuah kategori yang diberikan
  Discourse marker yang digunakan juga terbagi menjadi 5 jenis, yaitu justification (because, the reason being, dst.), explanation (in other words, for instance, quotesfor this reason(s), dst), deduction (as a consequence, in accordance with the above), rebuttal (despite, however, dst), dan conditionals (supposing that, in case that, dst)
- Frekuensi relatif dari setiap 6 tense dan 6 mood
- Kombinasi frekuensi relative dari setiap tense/mood
- Kemunculan setiap 6 tense dan 6 mood
- Tense, mood, dan kombinasi tense/mood yang paling sering muncul

Classifier yang digunakan adalah algoritma learning C4.5 decision tree.

Pendekatan menggunakan ontologi dilakukan oleh Ong, Litman, dan Brusilovsky (2014). Peneliti menggunakan 8 rule untuk mengidentifikasikan argumen dari setiap kalimat. Rule dibuat dengan menggunakan intuisi dari peneliti dan informal examination pada 9 esai dari keseluruhan esai yang ada. Korpus yang digunakan adalah 52 esai yang ditulis oleh mahasiswa pada dua mata kuliah psikologi di University of Pittsburgh. Selain menggunakan rule tersebut untuk mengkategorikan argumen pada esai, peneliti juga melakukan penilaian esai secara otomatis. Metode yang digunakan juga berupa algoritma rule-based yang dikembangkan juga dari intuisi peneliti namun dihubungkan dengan examination dari rubrik penilaian para ahli. Terdapat 5 rule yang kemudian digunakan untuk menghitung skor akhir dari esai.

Song, Heilman, Klebanov, dan Deane (2014) mencoba mengimplementasikan skema argumentasi untuk penilaian esai. Teori skema argumentasi yang digunakan berdasarkan pada teori Walton (1996) yang melibatkan beberapa penyesuaian di dalamnya. Data yang digunakan untuk protokol anotasi dalam menulis adalah analisa argumen dari ujian masuk graduate school. Level agreement antara manusia dengan esai yang dinilai oleh sistem masih di bawah nilai agreement antara manusia dengan manusia.

Masih minimnya korpus argumentasi yang dianotasi sehingga bisa digunakan sebagai batu pijakan untuk penelitian-penelitian lanjutan mendorong Stab dan Gurevych (2014) melakukan penelitian untuk menghasilkan anotasi komponen argumen beserta hubungan diantara komponen tersebut satu dengan yang lainnya. Anotasi manual dilakukan oleh 3 annotator pada 90 esai persuasif. Skema anotasi yang dibangun terdapat pada gambar 2. Output yang dihasilkan oleh penelitian ini adalah korpus esai serta anotasi yang sudah dilakukan pada esai tersebut mencakup identifikasi komponen argumen dan hubungannya.

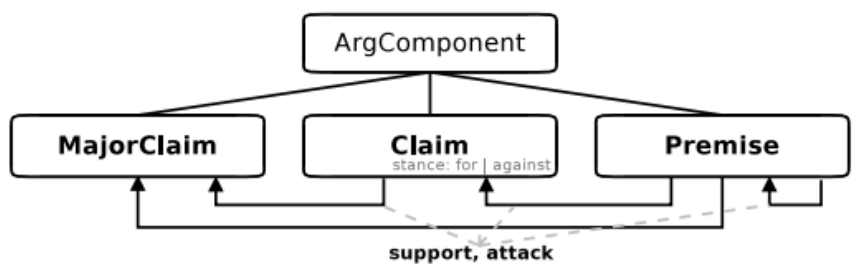

**Gambar 2. Skema Anotasi Argumen**

**(Stab dan Gurevych, 2014)**

Dari gambar 2, terlihat bahwa ada 3 macam komponen argumen, yaitu major claim, claim dan premise. Ketiga komponen ini saling berhubungan dimana premise akan mendukung (support) atau sebaliknya (attack) pada claim dan major claim. Komponen claim juga digunakan untuk mempertegas keberadaan major claim. Major claim dan claim tidak akan memiliki makna yang berarti sebagai sebuah argumentasi apabila tidak didukung oleh keberadaan premise.Di luar dari ketiga komponen tersebut, maka kalimat disebut sebagai non-argumen/none. Contoh dari beberapa kalimat yang termasuk dalam komponen argumen adalah sebagai berikut (diambil dari essay07 dari korpus Stab dan Gurevych, 2014):

1. Major Claim (MC)
   - *Newspapers have lost their competitive advantage to sustain their prolonged existence*
2. Claim (C)
   - *The print media has failed to keep its important role in the provision of information*
   - *The number of people reading newspapers may continue falling sharply, possibly leading to the close-downs of many in the coming time*
   - *Newspapers' production will have to face environmentalists on its way to be alive*
3. Premise (P)
   - *The internet has been more and more popular for recent years, providing people with a huge source of information*
   - *Contrary to the past when people had to wait long hours to take a daily newspaper, nowadays, they can acquire latest news updated every second through their mobile phones or computers connected to the internet, everywhere and at anytime*
4. None (N)
   - *As a result of this, print media such as newspapers have experienced a dramatic decline in the number of readers*
   - *The question arises as to whether or not a person spends an extra money buying newspapers to receive the same, even usually less information than those he can have with the internet?*

Tidak berhenti sampai pada anotasi argument pada 90 esai tersebut, Stab dan Gurevych (2014) memanfaatkan korpus yang sudah dianotasi tersebut untuk pekerjaan mengidentifikasi komponen argumen yang terklasifikasi menjadi 3 yaitu major claim, claim, dan premise. Satu label lain yaitu none digunakan untuk kalimat yang tidak memiliki komponen argumen sama sekali (non-argumen). Kemudian klasifikasi terhadap relasi argumen juga dilakukan. Klasifikasi dibagi menjadi dua yaitu support dan non-support (attack). Pada kedua task tersebut digunakan feature structural, lexical, syntactic, indicator dan contextual. Khusus untuk pendeteksian komponen argumen, Support Vector Machine (SVM) sebagai classifier yang hasilnya paling baik.

3. **Metodologi**

    Pada penelitian ini dilakukan pendekatan yang serupa untuk mendeteksi komponen argumen dari sekumpulan esai. Metode yang digunakan adalah memanfaatkan beberapa feature yang pernah dipakai yang kemudian dikombinasikan dengan beberapa feature dari penelitian yang berbeda.

    Korpus dari Stab dan Gurevych (2014) menyediakan data yang menunjukkan dimana saja lokasi dari argument component pada masing-masing esai. Korpus tersebut terdiri dari 90 esai. Dari esai yang sudah diberikan anotasi tersebut, dapat diperoleh daftar argument component. Sehingga, dapat dibangun list yang terdiri dari masing-masing argument component dari keseluruhan data. Kalimat yang tidak memuat argument component juga ditambahkan ke dalam list dengan pelabelan yang berbeda. Pada akhirnya, diperoleh sebuah list kalimat yang memuat seluruh argument component dan kalimat non-argumen.

    Selanjutnya, dilakukan feature extraction pada korpus yang terdiri dari 3 kategori, yaitu structural, lexical dan indicator. Secara umum, feature yang diujicobakan pada penelitian inimengambil sebagian dari yang dilakukan oleh Stab dan Gurevych (2014), Palau dan Moens (2009) serta mengadopsi daftar keyword yang ada padaKnott dan Dale (1993). Pada bagian akhir dilakukanperbandingan hasil apabila menggunakan daftar keyword dengan yang tidak menggunakan daftar keyword. Semua contoh pada penjelasan feature di bawah ini menggunakan data yang sama seperti yang dijelaskan pada bagian kajian literatur.

    - Structural feature

    Structural feature didefinisikan dengan menghitung beberapa komponen sebagai berikut:
    - Jumlah token dalam covering sentence.

        Covering sentence adalah kalimat yang memuat argument component. Jumlah token yang terletak pada covering sentence dihitung kemudian dijadikan feature.

        Contoh:

        *Hence, the print media has failed to keep its important role in the provision of information*

        Pada kalimat tersebut, komponen argumen sebagai claim dimulai setelah tanda koma hingga kata terakhir. Namun pada feature ini, semua token yang ada pada covering sentence dihitung tanpa terkecuali, sehingga outputnya adalah nilai 16.

- Boolean feature yang menunjukkan apakah argument component termuat secara utuh sebagai sebuah covering sentence atau tidak.

  Apabila argument component termuat di dalam covering sentence, maka nilai featurenya menjadi 1 (satu). Sebaliknya, apabila argument component tidak termuat secara utuh pada covering sentence maka nilai featurenya menjadi 0 (nol). Pada feature ini, nilai feature untuk kalimat non-argumen menjadi 0 (nol).

Contoh:

*The internet has been more and more popular for recent years, providing people with a huge source of information*

Kalimat di atas termasuk komponen argumen jenis premise dimana komponen argumen dimulai dari kata paling pertama hingga terakhir, hal ini berarti argument component termuat dalam covering sentence sehingga nilai feature pada kalimat tersebut adalah 1 (satu).

- Jumlah token dalam argument component.

  Secara sederhana feature ini hanya menghitung jumlah kata pada argument component. Untuk kalimat non-argumen, nilai feature adalah 0 (nol) karena tidak terdapat argument component di dalamnya.

  Contoh:

  *The number of people reading newspapers may continue falling sharply, possibly leading to the close-downs of many in the coming time*

  Kalimat di atas merupakan komponen argumen jenis claim. Feature ini menghitung jumlah kata pada komponen argumen, sehingga nilai feature pada kalimat tersebut adalah 21.

- Jumlah token sebelum dan sesudah argument component.

  Yang dimaksud adalah kata-kata pada covering sentence yang bukan bagian dari argument component. Misalnya jumlah token pada covering sentence adalah 20 dan jumlah token pada argument component adalah 15 maka nilai feature ini adalah 5. Untuk kalimat non-argumen, nilai feature nya adalah sama dengan jumlah token pada kalimat itu sendiri.

  Contoh:

  *Some people, however, still believe that they can exist for long time; others disagree, arguing that newspapers have lost their competitive advantage to sustain their prolonged existence*

  Komponen argumen jenis major claim pada kalimat di atas ada pada bagian "*newspapers have lost lost their competitive advantage to sustain their prolonged existence*" sehingga nilai feature pada kalimat ini adalah jumlah kata sebelum kata newspapers, yaitu 16.

- Jumlah tanda baca pada covering sentence.

  Semua bentuk tanda baca seperti titik, koma, titik koma, titik dua dan sebagainya dihitung jumlah kemunculannya untuk dijadikan nilai feature.

Contoh:

*Contrary to the past when people had to wait long hours to take a daily newspaper, nowadays, they can acquire latest news updated every second through their mobile phones or computers connected to the internet, everywhere and at anytime*

Jumlah tanda baca pada contoh di atas berjumlah 3, yaitu tanda koma (,), sehingga nilai featurenya adalah 3.

- Rasio perbandingan jumlah token pada argument component dengan jumlah token pada covering sentence.

  Jika jumlah token pada argument component adalah 15 dan jumlah token pada covering sentence adalah 20 maka nilai featurenya adalah pembagian antara kedua angka tersebut yaitu 0.75. Untuk kalimat non-argumen, nilainya adalah 0 (nol).

  Contoh:

  *The internet has been more and more popular for recent years, providing people with a huge source of information*

  Kalimat di atas secara keseluruhan adalah komponen argumen, maka nilai featurenya menjadi 19 dibagi 19, hasilnya adalah 1 (satu).

- Boolean feature yang menandakan apakah kalimat diakhiri dengan tanda tanya atau tidak.

  Apabila kalimat diakhiri dengan tanda tanya maka feature bernilai 1 (satu), jika tidak feature menjadi bernilai 0 (nol).

  Contoh:

  *The question arises as to whether or not a person spends an extra money buying newspapers to receive the same, even usually less information than those he can have with the internet?*

  Kalimat non-argumen diatas diakhiri dengan tanda tanya, sehingga nilai feature kalimat tersebut adalah 1 (satu)

- Lexical feature

    Lexical feature didefinisikan dengan menghitung beberapa komponen sebagai berikut:

    - Boolean feature kemunculan unigram

        Sebelum menentukan nilai feature, seluruh kemungkinan unigram di generate dari 90 esai. Hal ini akan membuat jumlah feature meningkat secara signifikan. Setelah diperoleh daftar unigram yang terbentuk, maka dilakukan lookup kepada semua kalimat, baik itu argumen atau non-argumen. Apabila unigram tersebut muncul pada kalimat maka nilai featurenya adalah 1 (satu), sebaliknya nilainya menjadi 0 (nol).

    - Boolean feature kumunculan bigram

        Sama seperti unigram, akan di generate semua kemungkinan bigram. Bigram adalah seluruh kemungkinan dua kata yang berurutan. Setelah diperoleh daftar bigram yang terbentuk, lookup pada seluruh kalimat argumen dan non-argumen dilakukan. Jika muncul maka nilainya adalah 1 (satu), jika tidak maka nilainya adalah 0 (nol).

    - Boolean feature kemunculan trigram

        Trigram merupakan seluruh kemungkinan tiga kata yang berurutan dari semua esai. Setelah dilakukan lookup sama seperti pada unigram dan bigram, jika terdapat kemunculan trigram dalam kalimat maka nilainya menjadi 1 (satu), jika tidak nilainya menjadi 0 (nol).

    - Boolean feature kemunculan modal

        Modal seperti should dan could seringkali muncul pada sebuah esai argumentasi untuk menunjukkan tingkat kepastian ketika mengekspresikan sebuah claim. Kemunculan modal tersebut dicek pada masing-masing kalimat sebagai Boolean feature.

        Contoh:

        *Newspapers' production will have to face environmentalists on its way to be alive*

        Karena pada kalimat di atas muncul kata *will* sebagai modal, maka nilai feature kalimat tersebut adalah 1 (satu).

- Indicator

    Daftar yang terdapat pada paper Knott dan Dale (1993) memuat 286 variasi keyword. Daftar keyword ini berperan seperti discourse marker. Beberapa diantaranya adalah *actually, by comparison, either, in this way, in conclusion*, dan sebagainya. Keyword tersebut dimanfaatkan sebagai feature. Feature ini dibagi menjadi dua jenis feature.

- Jumlah kemunculan keyword pada kalimat

  Apabila dalamkalimat (argumen/non-argumen) terdapat beberapa keyword yang muncul, maka jumlah itulah yang menjadi nilai feature.

- Boolean feature dari keyword

  Karena jumlah katanya berjumlah 286, maka terdapat 286 feature tambahan yang berupa Boolean feature. Feature ini hanya melihat apakah masing-masing keyword yang terdapat pada daftar keyword muncul pada kalimat atau tidak.

Setelah selesai melakukan feature extraction seperti rincian di atas, maka diperoleh kumpulan baris kalimat argumen/non-argumen yang memuat keseluruhan nilai feature. Pada parameter terakhir pada masing-masing kalimat ditambahkan satu feature penanda jenis dari kalimat tersebut apakah termasuk ke dalam major claim, claim, premise, atau none/non-argument. Gambaran data yang dihasilkan terlihat pada gambar 3.

$$\begin{pmatrix} & & & & & \\ & & & & & \\ & & & & & \\ & & & & & \\ f1 & f2 & f3 & f4 & \dots & fk \end{pmatrix} \begin{matrix} k1 \\ k2 \\ k3 \\ k4 \\ \dots \\ kn \end{matrix}$$

**Gambar 3. Ilustrasi data yang dihasilkan**

$f_1$ hingga $f_k$ menunjukkan feature pertama hingga feature ke-k. Nilai dari masing-masing feature bervariasi sesuai jenisnya, ada yang berupa jumlah, Boolean feature atau hasil dari operasi perhitungan. Feature terakhir merupakan penanda jenis kalimat. Nilainya hanya 0 (major claim), 1 (claim), 2 (premise) atau 3 (non-argumen). Jenis-jenis ini diperoleh dari korpus Stab dan Gurevych (2014).

$k_1$ hingga $k_n$ menunjukkan kalimat-kalimat yang dihasilkan dari ekstraksi korpus. Apabila merupakan kalimat argumen, maka yang masuk ke dalam k adalah argument component. Sedangkan apabila merupakan kalimat non-argumen, data yang masuk ke dalam k adalah kalimat non-argumen secara utuh.

4. **Hasil dan Pembahasan**

   Korpus yang digunakan merupakan korpus dari penelitian Stab dan Gurevych (2014). Seluruh esai dari korpus tersebut diekstrak menjadi kumpulan kalimat yang memiliki rincian berbagai jenis nilai feature. Jumlah esai yang digunakan adalah 90. Dari hasil ekstraksi korpus tersebut diperoleh instance (kalimat argumen dan non-argumen) sebanyak 1532. Instance tersebut terbagi menjadi 74 major claim, 348 claim, 867 premise, dan 243 none. Uji coba hasil feature extraction dilakukan dengan pengujian klasifikasi jenis argumen dengan menggunakan Weka data mining software. Kategori testing yang digunakan adalah 10-fold cross validation.

Untuk mengukur seberapa besar pengaruh penggunaan feature indicator pada feature secara keseluruhan, maka uji coba dibuat menjadi 2 jenis yaitu dengan penggunaan feature indicator tersebut dan tanpa menggunakan feature indicator. Seluruh klasifikasi dilakukan dengan menggunakan Support Vector Machine (SVM) sebagai classifier. Jumlah feature yang terlibat pada penelitian ini berjumlah 11978. Sedangkan, apabila feature indicator tidak digunakan, jumlah feature yang dihasilkan adalah 11749. Tabel rangkuman hasil dari ujicoba meliputi nilai akurasi dan waktu yang diperlukan untuk membangun model terdapat pada tabel 1. Sedangkan rincian klasifikasi terdapat pada tabel 2.

**Tabel 1. Akurasi klasifikasi dengan menggunakan SVM**

|  | **tanpa feature indicator** | **dengan feature indicator** |
| --- | --- | --- |
| 10-fold cross validation | 72.389% <br> (1109 terklasifikasi benar) <br> Waktu: 7.42 detik | 72.4543% <br> (1110 terklasifikasi benar) <br> Waktu: 8.11 detik |

**Tabel 2. Confusion Matrix**

*tanpa feature indicator*

| | | \multicolumn{4}{c}{*Hasil Prediksi*} |
| --- | --- | --- | --- | --- | --- |
| | | **Major Claim** | **Claim** | **Premise** | **None** |
| *Aktual* | **Major Claim** | 0 | 0 | 74 | 0 |
| | **Claim** | 0 | 0 | 346 | 2 |
| | **Premise** | 0 | 0 | 866 | 1 |
| | **None** | 0 | 0 | 0 | 243 |

*dengan feature indicator*

| | | \multicolumn{4}{c}{*Hasil Prediksi*} |
| --- | --- | --- | --- | --- | --- |
| | | **Major Claim** | **Claim** | **Premise** | **None** |
| *Aktual* | **Major Claim** | 0 | 0 | 74 | 0 |
| | **Claim** | 0 | 0 | 346 | 2 |
| | **Premise** | 0 | 0 | 867 | 0 |
| | **None** | 0 | 0 | 0 | 243 |

Technical ReportProgram Studi Doktor Ilmu Komputer
Fakultas Ilmu Komputer Universitas Indonesia, Juni 2015

Dari data yang diamati dari kedua tabel tersebut, terlihat bahwa pengklasifikasian yang salah terdapat pada major claim (MC) dan claim (C) yang menurut SVM masuk ke dalam kategori premise (P). Hanya ada 1 hingga 2 data dari yang seharusnya claim namun terklasifikasi sebagai none (N) dan juga yang seharusnya premise namun terklasifikasi sebagai none. Dari hasil pengamatan terhadap data di atas, terlihat bahwa feature yang dipakai belum mampu mewakili ciri dari major claim dan claim, sehingga perlu ditelusuri lebih lanjut mengenai bagaimana cara untuk mendeteksi major claim dan claim dengan benar. Hal ini senada dengan yang disimpulkan oleh Stab dan Gurevych (2014) yang mengatakan bahwa identifikasi claim dan major claim masih menghasilkan performance yang rendah.

Dari perbandingan data pada tabel, terdapat sedikit peningkatan akurasi ketika menggunakan feature indicator, yakni dari 72.369% menjadi 72.4543%. Hal ini berarti bahwa feature tersebut memiliki pengaruh meskipun tidak dengan signifikansi yang tinggi. Nilai akurasi tersebut masih lebih rendah dibandingkan yang dilakukan Stab dan Gurevych (2014) yaitu 77.3%. Nilai akurasi yang masih 72.4543% tersebut dikarenakan belum diimplementasikannya feature sintaktis, kontekstualdan word couple. Harapannya dengan penambahan feature tersebut, nilai akurasi akan menjadi lebih baik.

5. **Simpulan**

Simpulan yang dapat diperoleh dari penelitian yang dilakukan ini adalah sebagai berikut:
- Feature indicator dengan mengadopsi daftar keyword dari Knott dan Dale (1993) memiliki pengaruh pada klasifikasi komponen argumen meskipun tidak dengan signifikansi yang tinggi.
- Masih dibutuhkan feature yang bisa menjadi representasi dari komponen argumen major claim dan claim.

Beberapa perbaikan dan usulan yang bisa dikerjakan untuk penelitian lanjutan adalah:
- Menambahkan n-gram filtering dalam rangka memperkecil sparsity data pada feature n-gram.
- Menambahkan feature sintaktis, kontekstual dan beberapa feature lainnya supaya dapat memperkecil tingkat kesalahan klasifikasi
- Dilakukan mekanisme pendeteksian komponen argumen pada satu esai secara otomatis sehingga pekerjaan klasifikasi akan lebih dipermudah dan dinamis untuk berbagai jenis esai.

**Referensi**

1. Andreas Peldszus, and Manfred Stede. *From Argument Diagrams to Argumentation Mining in Texts: A Survey*. International Journal of Cognitive Informatics and Natural Intelligence (IJCINI), 7(1): 1-31, 2013

2. A. Knott and R. Dale. *Using Linguistic Phenomena to Motivate a Set of Rhetorical Relations*. Technical Report HCRC/RP-39, Edinburgh, Scotland, 1993.

3. Christian Stab, and Iryna Gurevych. *Annotating ArgumentComponents and Relations in Persuasive Essays*. Proceedings of COLING 2014, the 25th International Conference on Computational Linguistics: Technical Papers, pages 1501-1510, Dublin, Ireland, August 23-29, 2014.

4. Christian Stab, and Iryna Gurevych. *Identifying Argumentative Discourse Structures in Persuasive Essays*. Proceedings of the 2014 Conference on Empirical Methods in Natural Language Processing (EMNLP), pages 46-56, October 25-29, Doha, Qatar, 2014.

5. Douglas N. Walton. *Argumentation Schemes for Presumptive Reasoning*. Mahwah, NJ: Lawrence Erlbaum. 1996.

6. Eirini Florou, Stasinos Konstantopoulos, Antonis Kukurikos, and Pythagoras Karampiperis. *ArgumentExtraction for Supporting Public Policy Formulation*. Proceedings of the 7th Workshop on Language Technology for Cultural Heritage, Social Sciences, and Humanities, pages 49-54, Sofia, Bulgaria, August 8, 2013.

7. Jodie A. Butler and M. Anne Britt. *Investigating Instruction for Improving Revision of Argumentative Essays.* Written Communication, 28(1): 70-96, 2011.

8. Marie-Francine Moens, Erik Boiy, Raquel Mochales Palau, and Chris Reed. *Automatic Detection of Arguments in Legal Texts*. The 11th International Conference on Artificial Intelligence and Law, June 4-8, Stanford Law School, Stanford, California, 2007.

9. Nathan Ong, Diane Litman, and Alexandra Brusilovsky. *Ontology-Based ArgumentMining and Automatic Essay Scoring*. Proceedings of the First Workshop on Argumentation Mining, pages 24-28, Baltimore, Maryland USA, June 26, 2014.

10. Nitin Madnani, Michael Heilman, Joel Tetreault, and Martin Chodorow. *Identifying High-Level Organizational Elements in Argumentative Discourse*. Conference of the North American Chapter of the Association for Computational Linguistics: Human Language Technologies, pages 20-28, Montreal, Canada, June 3-8, 2012.

11. Raquel Mochales Palau, and Marie-Francine Moens. *Argumentation Mining: The Detection, Classification and Structure of Arguments in Text*. The 12th International Conference on Artificial Intelligence and Law, Barcelona. 2009

12. Yi Song, Michael Heilman, Beata Biegman Klebanov, and Paul Deane. *Applying Argumentation Schemes for Essay Scoring*. Proceedings of the First Workshop on Argumentation Mining, pages 69-78, Baltimore, Maryland USA, June 26, 2014.